\documentclass[letterpaper]{article} 
\usepackage{aaai25}  
\usepackage{times}  
\usepackage{helvet}  
\usepackage{courier}  
\usepackage[hyphens]{url}  
\usepackage{graphicx} 
\usepackage{natbib}  
\usepackage{caption} 
\usepackage{algorithm}
\usepackage{algorithmic}
\usepackage{bm}
\usepackage{amsmath}
\usepackage{amssymb}
\usepackage{multirow}
\usepackage{booktabs}
\usepackage{arydshln}
\usepackage{makecell}
\usepackage{tikz}
\usepackage{newfloat}
\usepackage{listings}
\DeclareCaptionStyle{ruled}{labelfont=normalfont,labelsep=colon,strut=off} 
\floatstyle{ruled}
\newfloat{listing}{tb}{lst}{}
\floatname{listing}{Listing}
\title{Improving Factuality in Large Language Models via Decoding-Time \\ Hallucinatory and Truthful Comparators}
\author{
    Dingkang Yang\textsuperscript{\rm 1}\equalcontrib\thanks{Done during the internship and implies the Ph.D. period end.},\, Dongling Xiao\textsuperscript{\rm 2}\equalcontrib,\, Jinjie Wei\textsuperscript{\rm 1},\, Mingcheng Li\textsuperscript{\rm 1},\, Zhaoyu Chen\textsuperscript{\rm 1}, \\
    Ke Li\textsuperscript{\rm 3}\thanks{Corresponding authors.},\, Lihua Zhang\textsuperscript{\rm 1\textdaggerdbl} \\
}
\affiliations{
    \textsuperscript{\rm 1}Academy for Engineering and Technology, Fudan University\\
    \textsuperscript{\rm 2}ByteDance\\
    \textsuperscript{\rm 3}Tencent Youtu Lab\\
     \{dkyang20, lihuazhang\}@fudan.edu.cn, xdluestc@outlook.com, tristanli@tencent.com

}

\usepackage{bibentry}

\begin{document}

\maketitle

\begin{abstract}
Despite their remarkable capabilities, Large Language Models (LLMs) are prone to generate responses that contradict verifiable facts, \textit{i.e.}, unfaithful hallucination content. 
Existing efforts generally focus on optimizing model parameters or editing semantic representations, which compromise the internal factual knowledge of target LLMs. 
In addition, hallucinations typically exhibit multifaceted patterns in downstream tasks, limiting the model's holistic performance across tasks.
In this paper, we propose a Comparator-driven Decoding-Time (CDT) framework to alleviate the response hallucination. Firstly, we construct hallucinatory and truthful comparators with multi-task fine-tuning samples. In this case, we present an instruction prototype-guided mixture of experts strategy to enhance the ability of the corresponding comparators to capture different hallucination or truthfulness patterns in distinct task instructions. CDT constrains next-token predictions to factuality-robust distributions by contrasting the logit differences between the target LLMs and these comparators. Systematic experiments on multiple downstream tasks show that our framework can significantly improve the model performance and response factuality.
The project will be released at \url{https://github.com/ydk122024/CDT}.

\end{abstract}

%

\section{Introduction}
Recently, the emergence of advanced Large Language Models (LLMs)~\cite{touvron2023llama-2,chatgpt,achiam2023gpt,jiang2023mistral} spearheads an understanding paradigm shift in the Natural Language Processing (NLP) and demonstrates superior capabilities in a wide range of tasks, such as translation, summarization, and dialog generation~\cite{see2017get,moon2019opendialkg}. Despite the impressive performance, LLMs typically generate seemingly plausible but non-factual claims, \textit{i.e.},  the hallucination dilemma. Hallucination content exhibits inconsistencies with real-world or user instructions, limiting the model credibility in realistic applications~\cite{huang2023survey}.

\begin{figure}[t]
  \centering
  \includegraphics[width=1.0\linewidth]{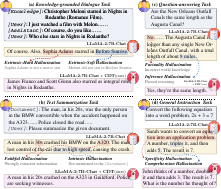}
  \caption{
  Illustration of multifaceted hallucination patterns generated by LLMs on different tasks ~\cite{das2023diving,zheng2023does}.
  }
\label{intro}
\end{figure}
The hallucination reasons are thought to be complex, such as biased corpora, maximum likelihood training objective, and blind instruction following~\cite{zhang2023siren}. All these factors potentially lead LLMs to assign non-zero probabilities to unfaithful token generation. 
Previous approaches have focused on two main aspects to provide solutions: optimizing model parameters and editing semantic representations. The former optimizes the model behavior directly from the response ranking via human preference alignment techniques~\cite{ouyang2022training,bai2022training,tian2023fine}, forcing the output to be more truthful. 
The Supervised Fine-tuning (SFT) is optionally used to mitigate task-specific hallucinations~\cite{jones2023teaching}.
However, secondary training is labor-intensive, and unintended knowledge infusion inadvertently encourages target LLMs to capture spurious generation patterns.
The latter works on regularizing and editing the hidden representations of LLMs to select the semantic spaces associated with correct statements~\cite{burns2022discovering,chen2024truth,li2024inference}.
Unfortunately, proactive representation activation may result in models hallucinating even knowing the correct semantics, interfering with internal factual knowledge~\cite{zou2023representation}.
More importantly, current LLMs suffer from multifaceted hallucination patterns that cause serious performance bottlenecks.
As shown in Figure~\ref{intro}, the model exhibits different types of hallucinations on each of the four tasks. There are even two hallucination patterns in the same response. For instance, \texttt{LLaMA2-7B-Chat} wrongly extracts information (\textit{i.e.}, faithful hallucination) in the text summarization and produces false content that is not present in the original document (\textit{i.e.}, intrinsic hallucination).
Previous studies have not resolved the coupled hallucination dilemma well.

Motivated by the above observations, we propose an efficient Comparator-driven Decoding-Time (CDT) framework to improve response factuality. The core philosophy is to equip target LLMs with comparators modeling different generative attributes separately during the decoding process, using logit distribution integration to facilitate next-token prediction in the factuality-robust directions.
We utilize explicit hallucinated/factual multi-task samples via Low-Rank Adaptation (LoRA)-based SFT to construct hallucinatory/truthful comparators based on the base models.
In this case, we design a hallucination perturbation adversarial mechanism to strengthen the factual knowledge mastery of the truthful comparator and mitigate the potential overfitting phenomena.
To assist target LLMs in eliminating the diverse hallucinations in cross-task generation, we present an instruction Prototype-guided Mixture of Experts (PME) strategy to empower the comparators with multi-task hallucination/truthfulness-aware capabilities. In the comparators, multiple LoRA adapters act as different experts to master different hallucination or truthfulness patterns through well-designed soft route gating.
CDT is capable of accomplishing effective attribute control without compromising the internal knowledge of target LLMs, maintaining the generated fluency and diversity.

The main contributions are summarized as follows:
\begin{itemize}
\item Our framework can help target LLMs remove non-factual knowledge from the output space during the decoding process in a product-integrated manner, significantly improving the robustness and factuality of model responses.
\item Our comparators are not limited to specific model structures and task types, in a plug-and-play form to guide the next-token prediction towards the hallucination-weak distribution. The proposed PME strategy offers the prospect and potential of eliminating hallucinations with multifaceted patterns in different tasks.
\item Extensive experiments are conducted on multiple NLP benchmarks. Comprehensive analyses show the broad applicability and effectiveness of our framework.
\end{itemize}

\section{Related Work}

\textbf{Hallucination Mitigation in LLMs.}
Hallucination in LLMs means the generated content is not supported by verifiable real-world facts or user instructions~\cite{huang2023survey,zhang2023siren}, leading to conflicts with input queries~\cite{rehman2023hallucination}, contextual information~\cite{shi2023trusting,wang2023survey}, and faithful semantics~\cite{bang2023multitask,hu2023large}.
Recently, extensive works have been presented aiming at mitigating the hallucination interference in LLMs across different dimensions, including retrieval-augmented generation~\cite{peng2023check,chern2023factool}, high-quality data construction~\cite{zhou2024lima,li2023textbooks}, and synthetic task migration~\cite{jones2023teaching}.
Among these, mainstream efforts have mainly focused on two aspects: optimizing model parameters and editing semantic representations.
The former argues that one of the most likely reasons for hallucinations is that the maximum-likelihood objective of language modeling leads LLMs to assign non-zero probabilities to unfaithful sentences outside the training data distribution~\cite{chuang2023dola}. To this end, these approaches promote harmless response generation through reinforcement learning from human feedback~\cite{ouyang2022training,bai2022training} and other preference alignment algorithms~\cite{tian2023fine,yang2024pediatricsgpt}.
For instance, \cite{tian2023fine} found that the Direct Preference Optimization (DPO)~\cite{rafailov2024direct} assists the model in aligning facticity from ranking.
The latter emphasizes activating attention heads or latent features into the truthful space through causal effects~\cite{li2024inference,chen2024truth,burns2022discovering,zhang2024truthx}.
For example, Truth Forest~\cite{chen2024truth} utilized multidimensional orthogonal probing to mine hidden true representations to improve model truthfulness.
Despite advances, current methods potentially undermine the inherent knowledge system of LLMs, causing performance bottlenecks on out-of-attention tasks.
In contrast, our framework guides out-of-the-box target LLMs during inference to enhance prediction factuality by mitigating hallucinations while improving truthfulness.

\noindent \textbf{Decoding-Time Intervention.}
Unlike secondary training for target LLMs, Decoding-time intervention was used earlier to enhance model fluency and coherence during inference by comparing expert models~\cite{li2022contrastive}. Previous studies found that controlling the output behavior at the decoding level contributed to avoiding toxic generated content~\cite{liu2021dexperts} and improving logical reasoning~\cite{wang2024chain}. These findings offer the prospect of refining response factuality through rational comparator formalization~\cite{zhang2023alleviating,shi2023trusting,chuang2023dola,kai2024sh2}.
For instance,
Dola~\cite{chuang2023dola} used the JSD divergence to measure the distribution of next-word predictions between the last and previous layers, and dynamically selected the previous layer that produces the largest difference to eliminate the hallucinatory semantics.
ICD~\cite{zhang2023alleviating} finetuned a factually weak base model during induce-the-contrast decoding as a penalty to guide LLMs more faithfully.
Nevertheless, these approaches usually fail to recognize multifaceted hallucination patterns in complex scenarios, leading to limitations and sensitivities in task types.
In comparison, the proposed CDT demonstrates adaptability and scalability over distinct tasks by hallucination/truthfulness-aware comparators.

\section{Methodology}

\begin{figure*}[t]
\centering
\includegraphics[width=\linewidth]{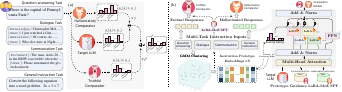}
  \caption{(a) Illustration of our Comparator-driven Decoding-Time (CDT) framework.
  The hallucinatory comparator gives higher weight to the incorrect \textit{Honolulu} while the truthful comparator favors the factual \textit{Harrisburg}, expressing the effective distribution control for the next-token prediction. The target LLM removes the hallucination through the CDT framework.
  (b) We construct hallucinatory/truthful comparators based on explicit hallucinated/factual instruction pairs via the LoRA-based SFT procedures. An instruction prototype-guided mixture of experts strategy is introduced during training to empower the comparators to help the target LLM improve response factuality in different downstream tasks.
  }
  \label{arc}
\end{figure*}

\subsection{Comparator-driven Decoding-Time Framework}
Figure~\ref{arc}(a) shows the high-level illustration of our CDT framework.
Given a target language model parameterized by $\theta$, the model generates a corresponding response $\bm{y}$ based on a query instruction $\bm{x}$ containing contextual information. The response is typically sampled from the probability distribution conditioned on $\bm{x}$. The autoregressive decoding process can be formulated as follows:
\begin{equation}
\small
y_{t} \sim p_{\theta} ( y_{t} \mid  \bm{x}, \bm{y}_{<t}) \propto \mathrm{exp\,logit_{\theta}}( y_{t} \mid  \bm{x}, \bm{y}_{<t}),
\end{equation}
where $y_{t}$ denotes the predicted token at time step $t$. $\bm{y}_{<t}$ contains the previously generated tokens across time step $t-1$.
With the maximum-likelihood-based training objective, the original decoding suffers from unavoidable hallucinations due to the model spontaneously assigning non-zero probabilities to unfaithful content~\cite{chuang2023dola}. From Figure~\ref{arc}(a), the model potentially relies on spurious correlations captured from the training data distribution to give false predictions.
To mitigate such issues, CDT aims to manipulate the next-token predictions of the target LLM into factuality-robust distributions by combining a hallucinatory comparator $\theta_{h}$, which generates text with non-factual attributes, and a truthful comparator $\theta_{f}$, which generates text with factual attributes.
These comparators come from fine-tuning the base version of the target LLM via explicit hallucinated/factual samples. The proposed CDT decoding is formulated as:
\begin{equation}
\begin{aligned}
\small
& y_{t} \sim p_{cdt} ( y_{t} \mid  \bm{x}, \bm{y}_{<t}) \\ 
& \propto p_{\theta} ( y_{t} \mid  \bm{x}, \bm{y}_{<t}) \frac{p_{\theta_{f}} ( y_{t} \mid  \bm{x}, \bm{y}_{<t})^{\beta }}{p_{\theta_{h}} ( y_{t} \mid  \bm{x}, \bm{y}_{<t})^{\gamma}},
\end{aligned}
\end{equation}
where $\beta$ and $\gamma$ are coefficients that control the degree of distribution modification.
Intuitively, the generated token will have a high probability when it has a high probability under $p_{\theta_{f}}$ and a low probability under $p_{\theta_{h}}$, facilitating factual prediction amplification and untruthful prediction trivialization.

In practice, we note that the output of the hallucinatory comparator is not always erroneous since it can still maintain basic linguistic common sense and grammar~\cite{li2022contrastive}. This means that indiscriminate trivialization potentially penalizes valuable aspects, leading to implausible tokens.
To address this issue, we introduce an adaptive plausibility constraint to penalize a subset $\mathcal{V}_{sub}$ of the output vocabulary $\mathcal{V}$ that correlates with the confidence level of the output distribution of the target LLMs:
\begin{equation}
\begin{aligned}
\small
 & \mathcal{V}_{sub}(\bm{y}_{<t}) = \{ y_{t} \in \mathcal{V}: \\
 & p_{\theta} ( y_{t} \mid  \bm{x}, \bm{y}_{<t})  \ge \delta \max_{w} p_{\theta} ( w \mid  \bm{x}, \bm{y}_{<t})  \}, \\
 & p_{cdt} ( y_{t} \mid  \bm{x}, \bm{y}_{<t}) = 0, \mathrm{if}\,\, y_{t} \notin \mathcal{V}_{sub} (\bm{y}_{<t}),
\end{aligned}
\end{equation}
where $\delta \in [0, 1]$ is a coefficient that controls the distribution truncation, with larger values indicating that only high-probability tokens are retained. The constraint ensures the response integrity of our framework while excluding the generation of untrustworthy tokens.

\subsection{Hallucinatory/Truthful Comparator Formalization}
To build behavior-controlled comparators, we activate hallucinatory/faithful mastery of the base models by the LoRA-based supervised fine-tuning (SFT)~\cite{hu2021lora}. The SFT dataset $\mathcal{D}$ contains multi-task instructions to empower the comparators to adapt the target LLM to improve factuality in different downstream tasks, including question-answering, knowledge-grounded dialogue, text summarization, and general tasks.
Specifically, given any input instruction $\bm{x} = (x_1, x_2, x_3, ...) \in \mathcal{D}$ and corresponding hallucinated/factual responses $\bm{y}^{h/f} = (y_1^{h/f}, y_2^{h/f}, y_3^{h/f}, ...) \in \mathcal{D}$,
the hallucinatory comparator can be optimized by the following fine-tuning objective:
\begin{equation}
\small
\mathcal{L}(\theta,\mathcal{D}) = \mathbb{E}_{(\bm{x},\bm{y}^{h})\sim \mathcal{D}}
\left [ -  {\textstyle \sum_{t=1}^{\left | \bm{y}^{h} \right | }}  \mathrm{log}\,p(y_{t}^{h} \mid \bm{x}, y_{<t}^{h}; \theta)  \right ].
\end{equation}
Although the truthful comparator can be optimized with the same objective,  we find that directly performing SFT via instruction pairs $(\bm{x}, \bm{y}^{f})$ results in overfitting, leading to the mundane faithfulness of the generated content.
Inspired by robustness learning~\cite{miyato2022adversarial}, we introduce a hallucination perturbation adversarial training mechanism to improve the training efficiency of the truthful comparator.
Specifically, we first backpropagate using the hallucinated responses to obtain the word embedding layer gradients. Then, we compute an adversarial perturbation based on the gradient and merge it into the original embedding to obtain the hallucination-aware adversarial samples. The perturbation is expressed as follows:
\begin{equation}
\small
\mathcal{P} = \varepsilon \cdot \bigtriangledown_{\bm{x}}\mathcal{L}(\theta, \bm{x}, \bm{y}^{h})  / \left \|\bigtriangledown_{\bm{x}}\mathcal{L}(\theta, \bm{x}, \bm{y}^{h}) \right \|_{2},
\end{equation}
where $\varepsilon$ and $\mathcal{L}(\cdot)$ represent the sign and loss functions, respectively.
The adversarial samples are utilized to backpropagate secondarily through the factual responses to obtain the cumulative adversarial gradients for model parameter updating.
Our strategy forces the model to progressively penalize hallucinated attributes from the perturbation during SFT, promoting more effective and factual semantic learning of the truthful comparator.
\subsection{Instruction Prototype-guided Mixture of Experts}

In the LoRA-based SFT phase of Figure~\ref{arc}(b), we present an instruction prototype-guided mixture of experts strategy to enhance the hallucination/truthfulness-aware capability of comparators.
Considering that instructions are the foundation for serving target LLMs to generate informative responses, the core philosophy of our strategy lies in learning prototype representations of multi-task instruction data with different probability density distributions via Gaussian Mixture Models~\cite{reynolds2009gaussian}.
Depending on the corresponding hallucinated or faithful response outputs, the prototype representations can guide multiple LoRA adapters acting as experts to master different hallucination or truthfulness patterns in supervised optimization, facilitating hallucinatory and truthful comparators to assist the target LLMs in generating factual content among diverse tasks during decoding time.
Specifically, 
given the multi-task instruction set $\left \{ \bm{x}_i  \right \}^{M}_{i=1}$ from the SFT dataset $\mathcal{D}$, the prototype representation acquisition is denoted as follows:
\begin{equation}
\small
\mathcal{C}_{[\bm{x}]} =\frac{{\textstyle \sum_{i=1}^{M}}\gamma_{i,k} \bm{h}_{x,i} } { {\textstyle \sum_{i=1}^{M} } \gamma_{i,k}},
\end{equation}
where $\bm{h}_{x,i}$ is the instruction feature corresponding to $\bm{x}_i$, which is extracted from the hidden state of the last layer in the target LLM and compressed and refined by principal component analysis for efficient computation.
$\gamma_{i,k}$ is the probability that each $\bm{h}_{x,i}$ belongs to the $k$-th Gaussian component, which is formulated as follows:
\begin{equation}
\small
\gamma_{i,k} = \frac{\pi_{k}\mathcal{N}(\bm{h}_{x,i}\mid\mu_{k}, {\textstyle \sum_{k}^{}} )}{ {\textstyle \sum_{j=1}^{K}}\pi_{j}\mathcal{N}(\bm{h}_{x,i}\mid\mu_{j}, \sum_{j} ) },
\end{equation}
where $\pi_{k}$, $\mu_{k}$, and $\sum_{k}$ are the mixture weight, the mean vector, and the covariance matrix of the $k$-th component, respectively. $\mathcal{N}(\cdot)$ denotes the probability density.

After that, the learned instruction prototypes are incorporated into the routing gating $G(\cdot)$ to activate the appropriate experts to capture speciﬁc hallucination/truthfulness patterns in any input $\bm{x}$. The gating vector is formulated as:
\begin{equation}
\small
  G(\bm{x
  })= \mathcal{S} \left ( \left ( (1-\mu) \odot \bm{x} + \\ \mu \odot \mathcal{C}_{[\bm{x}]} \bm{W}_{\mathcal{C}} \right )\bm{W}_{g} + \Re(\varphi (\bm{x}\bm{W}_{n}))\right ),
\end{equation}
where $\mathcal{S}(\cdot)$ and $\mu$ stand for the Softmax activation and trade-off coefficient, respectively.
In practice, we determine the index $k_{x} \leftarrow \arg\min_{k}d_{\mathcal{M}}(\bm{x}, \mathcal{C}_{[\bm{x}]})$ of the prototype $\mathcal{C}_{[\bm{x}]}$ to which $\bm{x}$ belongs via the Euclidean distance~\cite{danielsson1980euclidean}.
The distance is suitable for handling high-dimensional representations from multi-task SFT data with complex distributions since it is computationally efficient.
$\Re (\varphi (\bm{x}\bm{W}_{n}))$ is an additional noise term designed to regularize the load balancing among experts during the training process, where $\Re(\cdot)$ and $\varphi(\cdot)$ represent the standard normal distribution sampling and Softplus function, respectively. $\bm{W}_{\mathcal{C}}$, $\bm{W}_{g}$ and $\bm{W}_{n}$ are learnable parameters.
The forward process of obtaining the parameterized output $\bm{z}$ of all experts by integrating the soft route gating is expressed as follows:
\begin{equation}
\small
\bm{z} = \frac{\alpha}{r}{\sum_{n=1}^{N}} G(\bm{x})_n E_n(\bm{x}),
\end{equation}
where $r$ is the rank value and $\alpha$ is a constant hyper-parameter. $E_n(\cdot)$ represents the $n$-th LoRA expert.

\begin{table*}[t]
\renewcommand{\arraystretch}{1}
\centering
\resizebox{\linewidth}{!}{%
\begin{tabular}{cll|cccc|cccc}
\toprule
\multicolumn{3}{c|}{\multirow{2}{*}{\textbf{Methods}}} & \multicolumn{4}{c|}{\textbf{KNIGHT-Judge}}   & \multicolumn{4}{c}{\textbf{Alpaca-Judge}}                               \\ \Xcline{4-11}{0.4pt} 
\multicolumn{3}{c|}{}                         & \rule{0pt}{12pt} Accuracy (\%)  & Precision (\%) & Recall (\%)    & F1 Score (\%)  & Accuracy (\%)  & Precision (\%) & Recall (\%) & F1 Score (\%)  \\ \midrule
\multicolumn{3}{c|}{Baseline}                 & 45.13          & 43.90           & 62.50           & 51.58          & 43.68          & 45.05          & 81.33       & 59.27          \\
\multicolumn{3}{c|}{SFT}                      & 43.18          & 39.59          & 58.21          & 47.13          & 41.58          & 43.20           & 76.95       & 57.66          \\
\multicolumn{3}{c|}{FPO}                      & 44.42          & 41.67          & 61.04          & 49.53          & 43.72          & 45.46          & 82.63       & 59.14          \\
\multicolumn{3}{c|}{CD~\cite{li2022contrastive}}                       & 46.10           & 46.53          & \textbf{94.24} & 61.92          & 44.52          & 46.13          & 84.06       & 57.35          \\
\multicolumn{3}{c|}{ITI~\cite{li2024inference}}                      & 45.59          & 46.13          & 67.23          & 53.44          & 44.25          & 45.66          & 78.30        & 57.93          \\
\multicolumn{3}{c|}{SH2~\cite{kai2024sh2}}                      & 46.76          & 42.00             & 63.64          & 50.61          & 44.33          & 45.85          & 79.45       & 58.14          \\
\multicolumn{3}{c|}{DoLa~\cite{chuang2023dola}}                     & 46.62          & 41.81          & 65.79          & 52.25          & 44.67          & 46.27          & 84.94       & 58.91          \\ 
\multicolumn{3}{c|}{ICD~\cite{zhang2023alleviating}}                      & 47.24          & 47.76          & 90.32          & 63.93          & 45.00             & 47.38          & 85.33       & 60.15          \\
\midrule
\multicolumn{3}{c|}{\textbf{CDT} (Ours)}             & \textbf{50.16} & \textbf{50.60}  & 93.95          & \textbf{65.77} & \textbf{46.83} & \textbf{48.26} & \textbf{88.00} & \textbf{62.34} \\ \bottomrule
\end{tabular}
}
\caption{Comparison results on the KNIGHT-Judge and Alpaca-Judge datasets.}
\label{judge}
\end{table*}

\begin{table}[t]
\setlength{\tabcolsep}{10pt}
\renewcommand{\arraystretch}{1}
\centering
\resizebox{\linewidth}{!}{%
\begin{tabular}{cl|ccc}
\toprule
\multicolumn{2}{c|}{\multirow{2}{*}{\textbf{Methods}}} & \multicolumn{3}{c}{\textbf{TruthfulQA}} \\ \cline{3-5} 
\multicolumn{2}{c|}{}                         & \rule{0pt}{12pt} MC1 (\%)  & MC2 (\%)  & MC3 (\%)  \\ \midrule
\multicolumn{2}{c|}{Baseline}          & 37.62    & 54.60     & 28.12    \\
\multicolumn{2}{c|}{SFT}                      & 35.02    & 52.82    & 27.40     \\
\multicolumn{2}{c|}{FPO}                      & 37.98    & 55.34    & 29.65    \\
\multicolumn{2}{c|}{CD~\cite{li2022contrastive}}                       & 28.15    & 54.87    & 29.75    \\
\multicolumn{2}{c|}{ITI~\cite{li2024inference}}                      & 37.01    & 54.66    & 27.82    \\
\multicolumn{2}{c|}{SH2~\cite{li2024inference}}                      & 33.90     & 57.07    & 29.79    \\
\multicolumn{2}{c|}{DoLa~\cite{chuang2023dola}}                     & 32.97    & 60.84    & 29.50     \\
\multicolumn{2}{c|}{ICD~\cite{zhang2023alleviating}}                      & 46.32    & 69.08    & 41.25    \\
\midrule
\multicolumn{2}{c|}{\textbf{CDT} (Ours)}                      & \textbf{50.74}    & \textbf{75.96}    & \textbf{50.97}    \\ \bottomrule
\end{tabular}
}
\caption{Comparison results on the TruthfulQA dataset.}
\label{truthfulqa}
\end{table}

\section{Experiments}
\subsection{Datasets and Evaluation Metrics}
We construct extensive experiments on six datasets to evaluate different methods across multiple tasks.
Specifically, \textbf{KINGHT-Judge} is proposed to verify the model's ability to recognize hallucinations in the knowledge-driven dialog task.
The dataset contains the dialog contexts and factual responses from 616 samples sampled from the testing set of OpenDialKG~\cite{moon2019opendialkg}, as well as hallucinated responses generated by GPT-4~\cite{achiam2023gpt}.
During evaluations, the models are asked to classify factual/hallucinatory responses randomly sampled from each sample based on contexts to specify whether or not they contain hallucination information.
Similarly, we select 600 responses from HaluEval~\cite{li2023halueval} that are half-annotated as normal and hallucinatory against general user queries from Alpaca~\cite{taori2023stanford}.
These instructions constitute \textbf{Alpaca-Judge} to evaluate performance on general tasks. The evaluation metrics consist of Accuracy, Precision, Recall, and F1 Score.
\textbf{TruthfulQA}~\cite{lin2022truthfulqa} contains 817 questions across 38 categories to assess response truthfulness. We employ the multiple-choice task and measure the accuracy of LLMs in selecting answers from multiple correct/incorrect options via MC1, MC2, and MC3 metrics.
For text summarization, we extract 1,000 samples with harmless and safe content from vanilla \textbf{XSUM}~\cite{narayan2018don} to ensure the models perform rational summarization with given documents.
We use the ROUGE-1/2/L~\cite{lin2004rouge} to measure the generation quality. FactKB~\cite{feng2023factkb} and BERTScore~\cite{zhang2019bertscore} are utilized to evaluate the factual consistency of responses.
For open-ended generation tasks, we use factual responses in KNIGHT-Judge as Ground Truths (GTs) to construct \textbf{KNIGHT-Gen} to measure the generated content for given instructions. Moreover, \textbf{Alpaca-Gen} extracts 600 samples with normal responses as GTs to explore general generation following Alpaca-Judge. 
The FactKB, BERTScore, and DAE~\cite{goyal2020evaluating} are used to assess the predicted factuality.

\subsection{Implementation Details}
The main experiments are built on the \texttt{LLaMA-2} family in order to provide a fair comparison with existing methods. The model training is accomplished through the PyTorch platform using eight Nvidia A800 GPUs. 
To fine-tune the LLaMA2-base-driven comparators, the multi-task SFT dataset $\mathcal{D}$ is collected from HaluEval with 30K instructions having factual and hallucinated responses across NLP tasks. We also mix 10K held-out samples sampled from Alpaca and annotate the corresponding hallucinated responses.
The default number of instruction prototypes and LoRA experts are 32 and 4, respectively.
More detailed configurations can be found in the  \textbf{Appendix}.

\subsection{Comparison with State-of-the-Art Methods}
\textbf{Model Zoo.} 
We compare our CDT and reproducible methods. \textbf{Baseline}: \texttt{LLaMA2-7b-Chat} model~\cite{touvron2023llama-2}. \textbf{SFT}: The model is fine-tuned by factual multi-task instructions in $\mathcal{D}$. \textbf{FPO}: The model is optimized by DPO utilizing preference pairs from factual/hallucinated responses in $\mathcal{D}$.
Recent state-of-the-art (SOTA) works include CD~\cite{li2022contrastive}, ITI~\cite{li2024inference}, SH2~\cite{kai2024sh2}, DoLa~\cite{chuang2023dola}, and ICD~\cite{zhang2023alleviating}.

\noindent \textbf{Results on the KINGHT-Judge\&Alpaca-Judge.}
Table~\ref{judge} provides comparison results on the two datasets for hallucination recognition. We find that CDT achieves the best results on the vast majority of metrics, suggesting accurate judgments and understanding against potential hallucinations in the text. 
For instance, CDT brings 14.19\% and 3.07\% gains in terms of F1 scores on KINGHT-Judge and Alpaca-Judge for the baseline.
Conversely, the poor performance of SFT and FPO indicates that secondary training induces the models to learn spurious behavior cloning~\cite{schulman2023reinforcement} rather than factual patterns.
Our framework significantly outperforms prior SOTA ICD on KINGHT-Judge by 2.81\% average improvement, implying effective recognition of different extrinsic hallucinations.

\noindent \textbf{Results on the TruthfulQA.}
As shown in Table~\ref{truthfulqa}, our framework outperforms the decoding time methods ITI and ICD in all metrics on TruthfulQA. This implies that penalizing the coupled hallucinations only during decoding without considering the factual attributes that enhance the target LLMs is sub-optimal and inadequate.
Compared to ICD, the 9.5\%/9.9\%/23.6\% relative gains regarding MC1/MC2/MC3 on multi-category questions proves that CDT removes potential comprehension and factuality hallucinations~\cite{zheng2023does} in the question-answering task by giving correct answers with higher probabilities through the instruction prototype-guided mixture of experts strategy.

\begin{table}[t]
\setlength{\tabcolsep}{2pt}
\renewcommand{\arraystretch}{1}
\centering
\resizebox{\linewidth}{!}{
\begin{tabular}{cl|ccccc}
\toprule
\multicolumn{2}{c|}{\textbf{Methods}}       & R-1 & R-2 & R-L & FactKB & BERTScore \\ \midrule
\multicolumn{2}{c|}{Baseline}               & 17.22            & 3.55             & 13.29            & 38.79           & 82.50                 \\
\multicolumn{2}{c|}{SFT} & 17.58            & 3.61             & 13.75            & 41.85           & 81.03                \\
\multicolumn{2}{c|}{FPO}                    & 18.09            & 3.47             & 14.18            & 43.66           & 82.75                \\
\multicolumn{2}{c|}{CD~\cite{li2022contrastive}}                     & 19.55            & 3.77             & 14.61            & 67.30            & 83.31                \\
\multicolumn{2}{c|}{ITI~\cite{li2024inference}}                    & 17.94            & 3.68             & 15.45            & 57.94           & 82.87                \\
\multicolumn{2}{c|}{SH2~\cite{kai2024sh2}}                    & 19.75            & 3.83             & 14.73            & 71.22           & 83.92                \\
\multicolumn{2}{c|}{DoLa~\cite{chuang2023dola}}                   & 18.60             & 3.73             & 14.58            & 65.13           & 84.20                \\
\multicolumn{2}{c|}{ICD~\cite{zhang2023alleviating}}                    & 20.29            & 4.11             & \textbf{16.85}   & 70.26           & 84.06                \\ \midrule
\multicolumn{2}{c|}{\textbf{CDT} (Ours)}    & \textbf{21.11}   & \textbf{4.52}    & 16.33            & \textbf{73.66}  & \textbf{86.44}       \\ \bottomrule
\end{tabular}
}
\caption{Comparison results on the XSUM dataset. ``R-1/2/L'' means the ROUGE-1/2/L metrics.}
\label{xsum}
\end{table}

\begin{table*}[t]
\setlength{\tabcolsep}{12pt}
\renewcommand{\arraystretch}{0.9}
\centering
\resizebox{\linewidth}{!}{%
\begin{tabular}{cll|cccc|cccc}
\toprule
\multicolumn{3}{c|}{\multirow{2}{*}{\textbf{Methods}}} & \multicolumn{4}{c|}{\textbf{KNIGHT-Gen}}                                   & \multicolumn{4}{c}{\textbf{Alpaca-Gen}}                                   \\ \cline{4-11} 
\multicolumn{3}{c|}{}                         & \rule{0pt}{12pt} ROUGE-L        & BERTScore     & FactKB         & DAE            & ROUGE-L        & BERTScore     & FactKB         & DAE           \\ \midrule
\multicolumn{3}{c|}{Baseline}                 & 9.85           & 81.33          & 70.54          & 40.19          & 19.67          & 81.25          & 76.43          & 50.62          \\
\multicolumn{3}{c|}{SFT}                      & 9.92           & 81.14          & 71.43          & 39.76          & 20.54          & 81.23          & 77.25          & 50.11          \\
\multicolumn{3}{c|}{FPO}                      & 9.27           & 81.77          & 71.66          & 41.05          & 20.36          & 82.69          & 77.93          & 51.53          \\
\multicolumn{3}{c|}{CD~\cite{li2022contrastive}}                       & 10.80           & 82.35          & 72.27          & 42.83          & 22.42          & 82.07          & 78.20           & 52.00            \\
\multicolumn{3}{c|}{ITI~\cite{li2024inference}}                      & 11.26          & 82.59          & 74.43          & 42.46          & 23.56          & 83.31          & 80.17          & 52.29          \\
\multicolumn{3}{c|}{SH2~\cite{kai2024sh2}}                      & 11.13          & 83.57          & 72.68          & 44.10           & \textbf{24.35} & 83.99          & 83.02          & 55.43          \\
\multicolumn{3}{c|}{DoLa~\cite{chuang2023dola}}                     & 12.05          & 83.14          & 73.56          & 44.08          & 23.20           & 83.25          & 79.94          & 53.75          \\
\multicolumn{3}{c|}{ICD~\cite{zhang2023alleviating}}                      & 11.53          & 84.03          & 75.32          & 45.02          & 23.88          & 84.87          & 83.52          & 55.16          \\\midrule
\multicolumn{3}{c|}{\textbf{CDT} (Ours)}             & \textbf{12.16} & \textbf{85.47} & \textbf{77.42} & \textbf{47.39} & 23.90           & \textbf{86.06} & \textbf{85.82} & \textbf{56.50} \\ \bottomrule
\end{tabular}
}
\caption{Comparison results on the KNIGHT-Gen and Alpaca-Gen datasets.}
\label{gen}
\end{table*}

\begin{table}[t]
\setlength{\tabcolsep}{6pt}
\renewcommand{\arraystretch}{1}
\centering
\resizebox{\linewidth}{!}{%
\begin{tabular}{cl|cccc}
\toprule
\multicolumn{2}{c|}{\multirow{2}{*}{\textbf{Design}}}    & \begin{tabular}[c]{@{}c@{}}\textbf{KNIGHT-}\\ \textbf{Judge}\end{tabular} & \textbf{TruthfulQA}     & \textbf{XSUM}           & \begin{tabular}[c]{@{}c@{}}\textbf{Alpaca-}\\ \textbf{Gen}\end{tabular} \\ \cline{3-6} 
\multicolumn{2}{c|}{}                             & \rule{0pt}{10pt} F1 Score                                                   & MC1            & BERTScore      & DAE                                              \\ \midrule
\multicolumn{2}{c|}{\textbf{Full Framework}}               & \textbf{65.77}                                          & \textbf{50.74} & \textbf{86.44} & \textbf{56.50}                                         \\ \midrule
\multicolumn{6}{c}{\textbf{Importance of Comparators}}                                                                                                                                                        \\ \midrule
\multicolumn{2}{c|}{w/o HC} & 62.05                                                   & 39.53          & 84.97          & 52.12                                                 \\
\multicolumn{2}{c|}{+ Chat Version}  & 64.03                                                   & 48.36          & 85.15          & 55.74                                                 \\
\multicolumn{2}{c|}{w/o  TC}      & 54.61                                                   & 48.38          & 83.22          & 55.09                                                 \\
\multicolumn{2}{c|}{ + Chat Version}  & 65.75                                                   & 50.67          & \textbf{86.69} & \textbf{56.58}                                        \\ \midrule
\multicolumn{6}{c}{\textbf{Effectiveness of PME}}                                                                                                           \\ \midrule
\multicolumn{2}{c|}{w/o PME}                      & 62.94                                                   & 48.82          & 83.36          & 54.75                                                 \\
\multicolumn{2}{c|}{w/o  Prototypes}   & 64.06                                                   & 49.14          & 85.09          & 55.43                                                 \\ \midrule
\multicolumn{6}{c}{\textbf{Necessity of PAT}}                                                                                                                             \\ \midrule
\multicolumn{2}{c|}{w/o PAT}                      & 64.65                                                   & 49.57          & 85.64          & 55.96                                                 \\
\multicolumn{2}{c|}{ + PGD}           & 65.18                                                   & 49.83          & 85.91          & 56.20                                                  \\ \midrule
\multicolumn{6}{c}{\textbf{Impact of SFT Data Proportion}}                                                                                                                                                    \\ \midrule
\multicolumn{2}{c|}{w/o QA Data}                  & 64.40                                                    & 48.12          & 85.13          & 55.08                                                 \\
\multicolumn{2}{c|}{w/o Dialogue Data}            & 62.53                                                   & 48.93          & 85.38          & 54.71                                                 \\
\multicolumn{2}{c|}{w/o Summ. Data}       & 64.05                                                   & 49.27          & 84.45          & 54.66                                                 \\
\multicolumn{2}{c|}{w/o General Data}             & 65.19                                                   & 50.30           & 85.86          & 54.24                                                 \\ \bottomrule
\end{tabular}
}
\caption{Ablation study results on the different datasets. ``HC'' means the hallucinatory comparator. ``TC'' means the truthful comparator. ``+'' is the replacement operation. ``Summ.'' means the summarization. ``w/o'' is the without.}
\label{abl}
\end{table}

\noindent \textbf{Results on the XSUM.}
Table~\ref{xsum} demonstrates the text summarization capabilities of different models. According to the ROUGE metric, we find that the proposed framework generates content that covers the critical parts of the reference text more reasonably.
CDT assists the target baseline in improving 34.87\% and 3.94\% on FactKB and BERTScore, effectively enhancing response factuality and quality. In contrast, the improvement of the representation editing technique ITI is incremental due to enhancing truthfulness by intervening in the attention heads, which causes uninformative output and performance bottlenecks.

\begin{figure}[t]
  \centering
  \includegraphics[width=\linewidth]{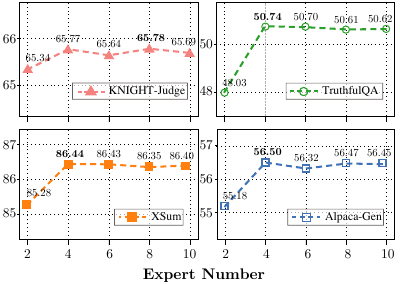}
  \caption{
 We show the effect of the number of experts on the performance of different tasks.
  }
\label{expert}
\end{figure}

\noindent \textbf{Results on the KINGHT-Gen\&Alpaca-Gen.}
We report the results of open-ended generation in Table~\ref{gen}. CDT improves response faithfulness on knowledge-grounded dialog and general-purpose tasks, consistently achieving the best results among three fact metrics.
Compared to other contrastive decoding methods, CDT has two advantages. First, unlike CD and DoLa, which emphasize probability differences between strong and weak models/layers, CDT enhances factuality in a more integrated way by enabling precise control of factuality-robust distributions of next-token predictions through comparators with different properties.
Also, our framework excels at capturing multiple hallucination types across tasks, leading to more effective hallucination mitigation during decoding time. 
The 6.39\% average gain across the fact metrics for the baseline confirms this.

\begin{figure*}[t]
\centering
\includegraphics[width=\linewidth]{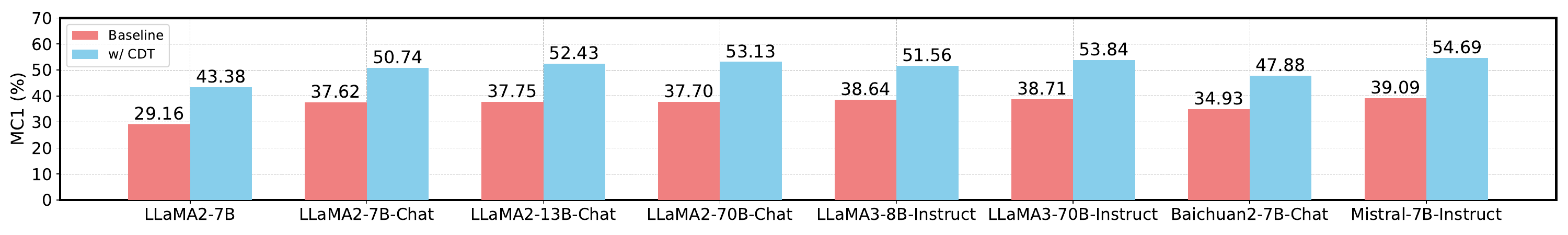}
  \caption{The extensibility analysis of our framework across different LLMs through the multiple-choice task on TruthfulQA.
  }
  \label{result}
\end{figure*}

\begin{figure}[t]
  \centering
  \includegraphics[width=\linewidth]{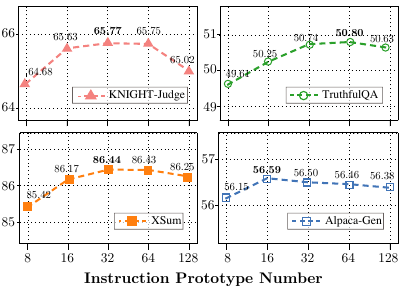}
  \caption{
 We show the effect of the number of instruction prototypes on the performance of different tasks.
  }
\label{Prototype}
\end{figure}

\subsection{Ablation Studies}
In Table~\ref{abl}, we perform systematic ablation studies to investigate the rationality of different designs in CDT.

\noindent \textbf{Importance of Different Comparators.}
We observe that the two comparators provide complementary contributions as performance deteriorates across different datasets when they are removed. Interestingly, the hallucinatory comparator excels in improving the discriminative and open-ended generation tasks, while the truthful comparator helps the target LLM to be more truthful in recognizing hallucinations and summarization demands.
When replacing the base model with the chat version, the truthful comparator has a minor improvement while it is ineffective for the hallucinatory comparator. 
This makes sense because the chat model undergoes RLHF with honesty as a key attribute, narrowing the gains from CDT contrasting logit differences.

\noindent \textbf{Effectiveness of Instruction Prototype-guided Mixture of Experts (PME).}
A single LoRA is used for the SFT process when there is no PME strategy in the comparators. The overall gain drops, which is more significant for the dialogue and summarization tasks. A plausible explanation is that PME facilitates the model to better capture sophisticated hallucinations in scenarios that require contextual understanding.
Moreover, instruction prototypes can guide LoRA experts to learn different hallucination/truthfulness patterns, bringing consistent gains across multiple tasks.

\noindent \textbf{Necessity of Perturbation Adversarial Training (PAT).}
We observe that the PAT mechanism plays an essential role because it prevents potential truthfulness learning trivialization and overfitting dilemmas in the truthful comparator training.
The intuitive justification is reflected in consistent gains across different datasets. Additionally, a PGD-based adversarial candidate~\cite{madry2017towards} is introduced to explore performance changes. The sub-optimal results from the candidate further verify the necessity of our mechanism.

\noindent \textbf{Impact of SFT Data Proportion.}
Different data for training the comparators are investigated by removing instruction pairs from each of the four parts. We discover that each type of SFT data is beneficial in reinforcing the benefits of the proposed framework. This trend suggests that sufficient multi-task data can effectively activate the comparators' ability to model different generation attributes.

\noindent \textbf{Effect of Expert Number.}
The number of LoRA experts determines the comparators' knowledge capacity to adapt to different tasks. In Figure~\ref{expert}, when only 2 experts are used, the model gives poorer results on XSUM and Alpaca-Gen. We conjecture that generative tasks typically require increased expert capacity to capture multifaceted hallucinations embedded in contextual information. The 4 experts achieve a trade-off between performance and training overhead. Excessively introducing experts can cause gain convergence.

\noindent \textbf{Effect of Instruction Prototype Number.}
Instruction prototypes guide the semantically relevant routing process to balance LoRA expert utilization. As shown in Figure~\ref{Prototype}, fewer prototypes (\textit{i.e.}, 8 and 16) perform poorly on the evaluation tasks. This stems from the inability of experts to master sufficient hallucination/truthfulness patterns, leading to performance bottlenecks. There is a considerable performance improvement on different tasks when we increase the number of prototypes, where 32 is the most appropriate. Conversely, too many prototypes are redundant as they exceed the pattern distribution boundaries~\cite{gou2023mixture}.

\subsection{Extensibility Analysis}
As a model-agnostic framework, we apply the CDT to 8 advanced LLMs spanning different sizes of the same family as well as different families of models. Figure~\ref{result} illustrates the results of the multiple-choice task on the hallucination perception dataset TruthfulQA.
In the \texttt{LLaMA} families, we find an upward trend in the benefits from CDT with increasing model version and size. The average improvement across the 6 vanilla baselines reaches 14.25\%. Moreover, CDT provides absolute gains of 12.95\% and 15.6\% for \texttt{Baichuan2-7B-Chat}~\cite{yang2023baichuan} and \texttt{Mistral-7B-Instruct}~\cite{jiang2023mistral}, respectively, significantly enhancing the corresponding truthfulness and factuality. These observations confirm the favorable extensibility and applicability of our framework.

\begin{table}[t]
\centering
\begin{tabular}{l|p{0.65\linewidth}} 
\toprule
\textbf{Methods}    & [\textit{Prompt}]: How long do diamonds last?                                                                              \\ \midrule
Baseline   & \textcolor{red}{Graphite} generally lasts between \textcolor{magenta}{10 and 15 million years} before it \textcolor{orange}{begins to weather and eventually decompose}.  \\ \midrule
ICD        & \textcolor{orange}{As the fossilized tree sap}, diamonds last between \textcolor{magenta}{5 and 8 billion years}.                                  \\ \midrule
\textbf{CDT} (Ours) & Typically, diamonds last between 1 and 4 billion years.                                                         \\ \bottomrule
\end{tabular}
\caption{Qualitative analysis regarding the generated responses of different models to the same prompt.
\textcolor{red}{Red}, \textcolor{magenta}{Magenta}, and \textcolor{orange}{Orange} texts represent faithful, factuality, and inference hallucinations, respectively.}
\label{vis}
\end{table}

\subsection{Qualitative Analysis}
In Table~\ref{vis}, we show the comparison results of generated contents for the same prompt by different models on the question-answering task.
The vanilla baseline yields multiple fabricated information across different hallucination patterns. Despite improvements in ICD, the hallucinations remained. Specifically, the claim that diamonds are inferred to be ``fossilized tree sap'' reflects the inference hallucination. ``5 and 8 billion years'' that conflict with world knowledge reflects the factuality hallucination. In contrast, CDT effectively corrects the multifaceted misinformation, thereby demonstrating the effectiveness of our framework.

\section{Conclusion}
In this paper, we introduce CDT, a decoding-time framework to efficiently remove the multifaceted hallucination dilemma of target LLMs via hallucinatory and truthful Comparators. We devise an instruction prototype-guided mixture of experts strategy to empower comparators with different hallucination/truthfulness-aware mastery. CDT facilitates more factual model responses in multiple downstream tasks by controlling the next-token prediction.

\section{Acknowledgments}
This work is supported in part by the National Key R\&D Program of China (2021ZD0113502).
 
\bibliography{aaai25}

\section{Appendix}
\subsection{Training Configuration Details}

\textbf{Comparator Training}. The overall model training is accomplished through the PyTorch platform with Accelerate and DeepSpeed packages using eight Nvidia A800 GPUs.
The AdamW optimizer~\cite{loshchilov2017decoupled} is adopted for network optimization
To obtain hallucinatory and truthful comparators, we perform the LoRA-based SFT procedures based on the base models (\textit{i.e.}, \texttt{LLaMA2-7B-base}). The rank value $r$ and constant hyper-parameter $\alpha$ are set to 8 and 32, respectively. The fine-tuned LoRA targets include $q_{proj}$, $k_{proj}$, and $v_{proj}$.
The gradient accumulation step is 4, resulting in 256 batch sizes. The maximum cutoff length and learning rate are 1024 and 5e-5, respectively. For the hallucinatory comparator, we train 3 epochs to ensure the effective hallucination generation capability of the model. For the truthful comparator, we take the proposed hallucination perturbation adversarial training to be completed with 2 epochs.
In the PME strategy-based route gating, the moderation factor $\mu$ is set to 0.1 to trade off the informativeness between the vanilla token and the corresponding instruction prototype.

\noindent \textbf{Prototype Acquisition}.
In order to obtain instruction prototypes with consistent semantic priors, we extract preliminary instruction features from the hidden state of the last layer of the target LLM with a dimension of 4096. For efficient prototype acquisition via the GMM algorithm~\cite{reynolds2009gaussian}, we use principal component analysis to compress and refine the feature dimension to 512.

\noindent \textbf{Inference Configuration}.
In Table~\ref{configurations}, we show the inference parameter configurations of the models under the CDT framework on different datasets.
The optimal hyperparameters are determined by the grid search.

\begin{table}[h]
\centering
\resizebox{0.85\linewidth}{!}{%
\begin{tabular}{c|cccc}
\toprule
Datasets     & $\beta$   & $\gamma$   & $\delta$    & Temperature \\ \midrule
KNIGHT-Judge & 0.5 & 0.5  & 0.1  & 0.5         \\
Alpaca-Judge & 0.5 & 0.5  & 0.1  & 0.05        \\
TruthfulQA   & 0.1 & 2    & 0    & 0.95        \\
XSUM         & 2   & 0.01 & 0.05 & 0.95        \\
KNIGHT-Gen   & 1   & 1    & 0.1  & 0.95        \\
Alpaca-Gen   & 1   & 2    & 0.1  & 0.95        \\ \bottomrule
\end{tabular}
}
\caption{Inference parameter configurations for models in CDT on different datasets.}
\label{configurations}
\vspace{-8pt}
\end{table}

\subsection{SFT Dataset Acquisition}

We carefully formulate the multi-task SFT dataset in order to train adequate hallucinatory and truthful comparators.
For the hallucinatory comparator,
we use 30k hallucinated instruction pairs from HaluEval, which contain multifaceted hallucinatory patterns across the three tasks.
These tasks contain question-answering, knowledge-grounded dialogue, and text summarization tasks.
We also mix 10K held-out samples sampled from Alpaca and annotate the corresponding hallucinated responses.
These samples and test data are not overlapped to avoid data leakage.
We prompt GPT-4 to generate hallucinated responses in a role-playing manner, which encompasses three hallucination patterns: unverifiable, non-factual, and irrelevant hallucinations.
The system prompt template is displayed in Table~\ref{hallucinated_prompt}.

For the truthful comparator, we utilize the same instructions used to train the hallucinatory comparator but with factual responses. We find that vanilla responses in the question-answering task are usually correct but concise, impacting model training and adaptability on generative tasks. 
Based on these observations, we guide GPT-4 through a self-instruct paradigm using the vanilla responses as seed instructions for factuality-enhanced response expansion. After that, we perform rigorous manual checks to eliminate potential noise and misleading information from the expanded responses. The prompt template is displayed in Table~\ref{fact_prompt}.

\begin{table}[t]
\centering
\begin{tabular}{lp{0.5\linewidth}}
\toprule
\textit{Prompt for Generating Hallucinated Responses }                       \\ \midrule
\begin{tabular}[c]{@{}p{0.95\linewidth}@{}}{[}\textit{INST}{]}\textless{}\textit{SYS}\textgreater\\ You are a proficient hallucination generator. Given the ``Instruction'', the ``Input'', and the ``Factual Response'', your goal is to write a corresponding hallucinated response that seems plausible but is incorrect. The hallucinated response you write can follow one of three hallucinatory patterns, including unverifiable, non-factual, or irrelevant information.\textless{}/\textit{SYS}\textgreater\\ \textless{}\textit{USER}\textgreater\\ ``Instruction'': \{instruction\}\\ ``Input'': \{input\}\\ ``Factual Response'':\{factual response\}\\ \textless{}/\textit{USER}\textgreater{}{[}\textit{INST}{]}\end{tabular} \\ \bottomrule
\end{tabular}
\caption{The system prompt template for generating hallucinated responses.}
\label{hallucinated_prompt}
\vspace{-8pt}
\end{table}

\begin{table}[t]
\centering
\begin{tabular}{lp{0.5\linewidth}}
\toprule
\textit{Prompt for Expanding Factual Responses}                        \\ \midrule
\begin{tabular}[c]{@{}p{0.95\linewidth}@{}}{[}\textit{INST}{]}\textless{}\textit{SYS}\textgreater\\ You are a helpful and faithful assistant. Please help me with the rephrasing task below. Don't share false information if you recognize unverifiable and unknown content.\textless{}/\textit{SYS}\textgreater\\ \textless{}\textit{USER}\textgreater\\
Please provide the extended rephrasing of the entities or short instances in the “Output”, taking into account the contextual information in the “Instruction”:. The reformulation must be informative and logical in terms of linguistic expression while following the meaning of the vanilla Output. Please output the rephrased content directly.
\\ ``Instruction'': \{instruction\}\\ ``Output'': \{output\}\\ \textless{}/\textit{USER}\textgreater{}{[}\textit{INST}{]}\end{tabular} \\ \bottomrule
\end{tabular}
\caption{The system prompt template for expanding factual responses.}
\label{fact_prompt}
\vspace{-8pt}
\end{table}

\begin{table*}[t]
\begin{tabular}{l|p{0.85\linewidth}}
\toprule
\textbf{Methods}    & \begin{tabular}[c]{@{}p{\linewidth}@{}}{[}\textit{Knowledge}{]}: The Usual Suspects is written by Christopher McQuarrie. Christopher McQuarrie wrote Valkyrie. Valkyrie has genre Thriller. Valkyrie is starring Tom Wilkinson.\\ {[}\textit{Human}{]}: Do you like : The Usual Suspects?\\ {[}\textit{Assistant}{]}: I have never seen it. I know it was written by Christopher McQuarrie. \\ {[}\textit{Human}{]}: Valkyrie also is from him. It is Thriller genre. Do you like thrillers?\end{tabular} \\ \midrule
Baseline   & Yes, I have seen \textcolor{brown}{Valhalla Valkyrie} with \textcolor{purple}{James Bond} in it.                                                                                                                                                                                                                                                                            \\ \midrule
ICD        & Yes, I like thrillers. I have seen \textcolor{purple}{Keira Knightley} in it.                                                                                                                                                                                                                                                                                                                                                                                              \\ \midrule
\textbf{CDT}  (Ours) & Yes, I do. Valhalla was released in 2008. I have seen Tom Wilkinson play a key role in it.                                                                                                                                                                                                                                                                                                                                                              \\ \bottomrule
\end{tabular}
\caption{Qualitative analysis regarding the generated responses of different models on the knowledge-grounded dialogue task.
\textcolor{brown}{Brown} and \textcolor{purple}{Purple} texts represent extrinsic-hard and extrinsic-soft hallucinations, respectively.}
\label{vis_dialog}
\end{table*}

\begin{table*}[t]
\begin{tabular}{l|p{0.85\linewidth}}
\toprule
\textbf{Methods}    & \begin{tabular}[c]{@{}p{\linewidth}@{}}{[}\textit{Document}{]}: Leigh-Anne Mahachi, 22, was discovered  at a property in Spotswood Close, in Gleadless Valley, on Tuesday, but died later in hospital. A post-mortem examination found she died from multiple stab wounds. South Yorkshire Police described her death as a ``targeted attack''. A 37-year-old man from Colchester, Essex, arrested on suspicion of murder on Tuesday, remains in police custody.
\\ {[}\textit{User}{]}: Please summarize the given document.\end{tabular} \\ \midrule
Baseline   & Leigh-Anne Mahachi, 22, died from a \textcolor{teal}{one-off stabbing} in the Gleadless Valley. The killer was a 37-year-old man known for \textcolor{violet}{his past violent behavior}.
           \\ \midrule
ICD        & Leigh-Anne Mahachi, 22, \textcolor{teal}{was whacked to death} in Headless Valley. A 37-year-old man has been arrested on suspicion of murder.
  \\ \midrule
\textbf{CDT} (Ours)& Leigh-Anne Mahachi, 22, died of multiple stab wounds in Headless Valley. A 37-year-old male suspect is in police custody.
  \\ \bottomrule
\end{tabular}
\caption{Qualitative analysis regarding the generated responses of different models on the text summarization task.
\textcolor{teal}{Teal} and \textcolor{violet}{Violet} texts represent faithful and intrinsic hallucinations, respectively.}
\label{vis_summ}
\end{table*}

\begin{table}[t]
\setlength{\tabcolsep}{3pt}
\renewcommand{\arraystretch}{1.1}
\centering
\resizebox{\linewidth}{!}{%
\begin{tabular}{cl|cccc}
\toprule
\multicolumn{2}{c|}{\multirow{2}{*}{\textbf{Design}}}    & \begin{tabular}[c]{@{}c@{}}\textbf{KNIGHT-}\\ \textbf{Judge}\end{tabular} & \textbf{TruthfulQA}     & \textbf{XSUM}           & \begin{tabular}[c]{@{}c@{}}\textbf{Alpaca-}\\ \textbf{Gen}\end{tabular} \\ \cline{3-6} 
\multicolumn{2}{c|}{}                             & \rule{0pt}{10pt} F1 Score                                                   & MC1            & BERTScore      & DAE                                              \\ \midrule
\multicolumn{2}{c|}{\textbf{Full Framework}}               & \textbf{65.77}                                          & \textbf{50.74} & \textbf{86.44} & \textbf{56.50}                                         \\ \midrule
\multicolumn{6}{c}{\textbf{Impact of Prototype Learning}}                                                                                                                                                        \\ \midrule
\multicolumn{2}{c|}{w/ K-Means++} & 64.53                                                   & 49.35          & 86.09          & 55.97                                                \\
\multicolumn{2}{c|}{w/ Mean-Shift}  & 64.81                                                   & 49.89          & 86.23          & 56.34                                                \\
 \midrule
\multicolumn{6}{c}{\textbf{Necessity of Adaptive Plausibility Constraint (APC)}}                                                                                                           \\ \midrule
\multicolumn{2}{c|}{w/o APC}                      & 65.20                                                   & \textbf{50.74}          & 86.05          & 56.12                                                 \\
\bottomrule
\end{tabular}
}
\caption{Ablation study results on the different datasets. ``w/o'' and ``w/'' represent the without and with, respectively.}
\label{abl_app}
\vspace{-8pt}
\end{table}

\subsection{Model Response Comparison}
We further show the qualitative comparison results of the different methods for generating responses. As shown in Table~\ref{vis_dialog}, the vanilla baseline has poor instruction following ability on the knowledge-grounded dialogue task, generating two types of hallucinations. Specifically, the baseline incorrectly recognizes ``Valhalla Valkyrie'' as a movie and believes that ``James Bond'' starred in it. Additionally, the previous ICD, while avoiding the extrinsic-hard hallucination, still generated the incongruous actor named ``Keira Knightley'', suffering from the extrinsic-soft hallucination interference. In contrast, the proposed CDT effectively removes all potential hallucination patterns, giving a more authentic and informative response.

In Table~\ref{vis_summ}, we explore the capabilities of different methods on the text summarization task. We find that the baseline model gives summarization content that is inconsistent with the original document (\textit{i.e.}, summarizing ``multiple stabs'' as ``one-off stabbing''), causing the faithful hallucination. At the same time, the model also fabricates content that is not in the original document due to the intrinsic hallucination (\textit{i.e.}, his past violent behavior). Although ICD mitigates the intrinsic hallucination, it still erroneously summarizes ``multiple stab wounds'' as ``whack death''. In comparison, our framework effectively corrects the misinformation produced by the baseline, confirming its effectiveness and necessity.

\subsection{Additional Ablation Studies}
From Table~\ref{abl_app}, we provide more ablation studies to investigate the effects of prototype learning and adaptive plausibility constraint (APC) strategies on performance.

\noindent \textbf{Impact of Prototype Learning}.
Here, we replace the default GMM method with candidate K-Means++~\cite{bahmani2012scalable} and Mean-Shift~\cite{comaniciu2002mean} algorithms to perform instruction prototype learning.
The consistently lower gains of the candidates across the four tasks demonstrate the reasonability of the default setting. 
The advantage derives from the fact that GMM assumes that the data is generated from a mixture of multiple Gaussian distributions, which is more robust to complex multi-task instruction data distributions.
Meanwhile, this method can handle ambiguous boundaries well when the data points belong to more than one prototype since it provides the corresponding probabilistic confidence estimates. On the contrary, K-means++ and Mean-Shift belong to hard clustering algorithms, which are susceptible to noise and additional parameter choices.

\begin{figure}[t]
  \centering
  \includegraphics[width=\linewidth]{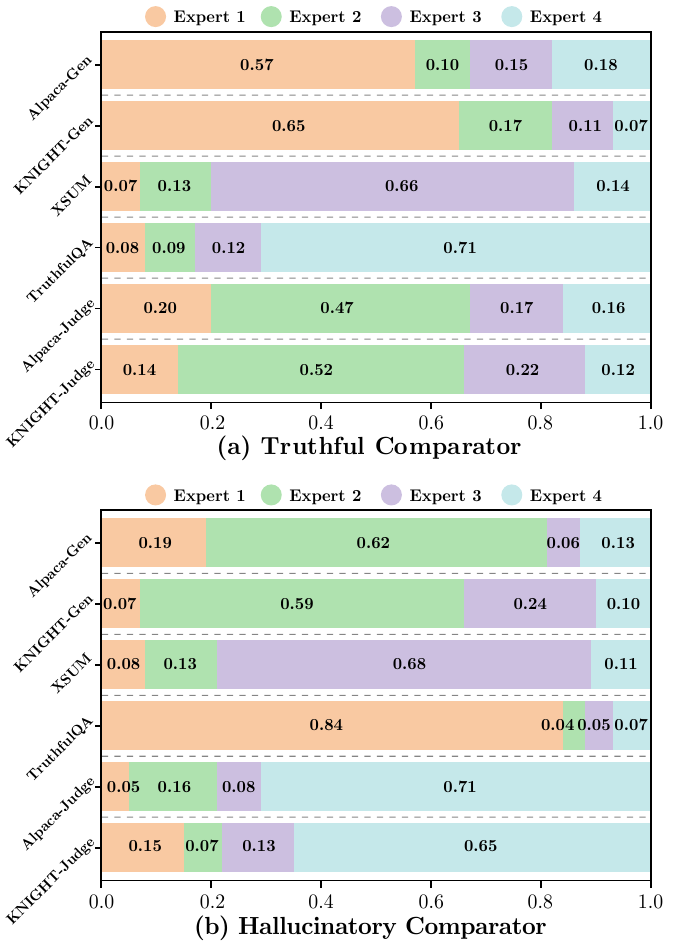}
  \caption{
We show the visualization results of expert routing selection on different task data.
  }
\label{expert_dis}
\end{figure}

\noindent \textbf{Necessity of Adaptive Plausibility Constraint}.
We observe that implementing the adaptive plausibility constraint provides consistent performance improvements in text generation scenarios. This means penalizing a subset of candidates from the vocabulary avoids compromising basic linguistic common sense and grammar, and maintains the logicality of the model response. 
In addition, the results on TruthfulQA are not affected since we assign scores based on probability without using the APC mechanism in the default setting. 

\begin{figure}[t]
  \centering
  \includegraphics[width=0.85\linewidth]{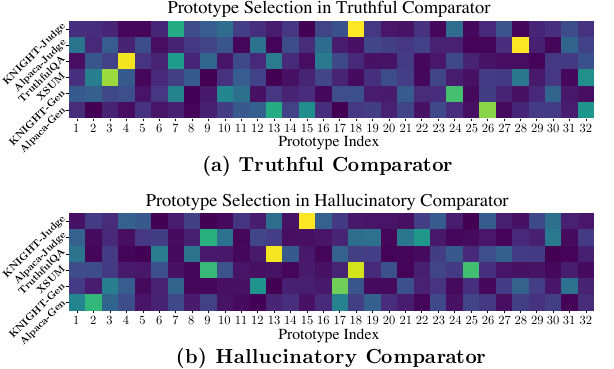}
  \caption{
We show the visualization results of instruction prototype selection on different task data. 1 through 32 on the horizontal coordinate represent prototype indexes.}
\label{Prototype_Selection}
\vspace{-5pt}
\end{figure}

\subsection{Expert Routing Selection}
To confirm the duties of different LoRA experts in the routing selection, we visualize the normalized routing weights from 4 experts on 6 tasks in Figure~\ref{expert_dis}(a)\&(b). The important observations are as follows. Although there are differences in expert utilization in different comparators, they all exhibit specialization for specific types of tasks.
For instance, Expert 1 in the truthful comparator excels in the open-ended generation tasks (\textit{i.e.}, Alpaca-Gen and KNIGHT-Gen) due to being activated emphatically. Expert 4 acquires truthfulness patterns across different instructions on the question-answering task (\textit{i.e.}, TruthfulQA) compared to the other experts. 
Conversely, Expert 4 in the hallucinatory comparator is more proficient in hallucination patterns on the hallucination judgment task, resulting in higher activation levels on the Alpaca-Judge and KNIGHT-Judge datasets.
Furthermore, Expert 3 favors handling the text summarization task in both comparators. These findings further confirm the effectiveness of the PME strategy and the advantages of CDT in different downstream tasks.

\subsection{Instruction Prototype Selection}
Figure~\ref{Prototype_Selection}(a)\&(b) provides the visualization results of the normalized probabilities of instruction prototype selection. We observe that truthful/hallucinatory comparators can enhance multifaceted truthfulness/hallucination-aware capabilities by selecting appropriate prototypes when dealing with different tasks.
For example, the truthful comparator learns factuality and inference patterns in TruthfulQA primarily through prototypes 4 and 7, reinforcing the truthfulness of CDT's output. In contrast, the hallucinatory comparator typically uses prototypes 9, 18, and 25 to capture multifaceted hallucination patterns (\textit{e.g.}, faithful and intrinsic hallucinations) in XSUM, guiding CDT to weaken erroneous information generation.

\subsection{Limitations}
Our framework will slightly increase the inference time overhead due to requiring forward propagation twice. Additionally, CDT cannot create new knowledge out of thin air, leading to potential bottlenecks when meeting situations outside of internal knowledge.
In future work, we will equip efficient decoding mechanisms and external tools to alleviate these limitations.

\end{document}